\documentclass[letterpaper,10pt,conference]{ieeeconf}

\IEEEoverridecommandlockouts

\usepackage[utf8]{inputenc} % allow utf-8 input
\usepackage[T1]{fontenc}    % use 8-bit T1 fonts
\usepackage{hyperref}       % hyperlinks
\usepackage{url}            % simple URL typesetting
\usepackage{booktabs}       % professional-quality tables
\usepackage{amsfonts}       % blackboard math symbols
\usepackage{nicefrac}       % compact symbols for 1/2, etc.
\usepackage{microtype}      % microtypography

\usepackage{times}
\usepackage{graphicx} % more modern
\usepackage{caption}
\usepackage{subcaption}
\usepackage{wrapfig}

\usepackage{algorithm}
\usepackage{algorithmic}

\usepackage{hyperref}

\usepackage{amsmath}
\usepackage{amssymb}

\usepackage[font=footnotesize]{caption,subcaption}
\usepackage{sidecap}

\usepackage{xspace}

\usepackage{color}

\newtheorem{theorem}{Theorem}[section]

\newcommand{\bx}{\mathbf{x}}
\newcommand{\bu}{\mathbf{u}}
\newcommand{\bo}{\mathbf{o}}

\newcommand{\dkl}{{D_{\textsc{kl}}}}
\newcommand{\dklmax}{{D_{\textsc{kl}}^{\max}}}
\newcommand{\dtv}{{D_{\textsc{TV}}}}

\newcommand{\pimix}{{\pi_{\textsc{mix}}}}

\newcommand{\picoach}{{\pi_{\textsc{coach}}}}
\newcommand{\pith}{{\pi_\theta}}
\newcommand{\pila}{{\pi_\lambda}}
\newcommand{\pilat}{{\pi_\lambda^t}}
\newcommand{\epsts}{{\epsilon_{\text{\tiny{$\theta *$}}}}}
\newcommand{\epslt}{\epsilon_{\text{\tiny{$\lambda \theta$}}}}

\newcommand{\E}{\mathbb{E}}
\newcommand{\tr}{\text{tr}}

\newcommand{\plato}{PLATO\xspace}

\newcommand{\pushright}[1]{\ifmeasuring@#1\else\omit\hfill$\displaystyle#1$\fi\ignorespaces}

\newcommand{\specialcell}[2][c]{%
  \begin{tabular}[#1]{@{}c@{}}#2\end{tabular}}
  
\DeclareGraphicsExtensions{.png}
\graphicspath{ {./figures/} }

\newcommand{\NEW}[1]{#1}

\title{\LARGE \bf
PLATO: Policy Learning using Adaptive Trajectory Optimization
}

\author{Gregory Kahn$^{1}$, Tianhao Zhang$^{1}$, Sergey Levine$^{1}$, Pieter Abbeel$^{1,2,3}$%
\thanks{$^{1}$Berkeley AI Research (BAIR), University of California, Berkeley}%
\thanks{$^{2}$OpenAI}%
\thanks{$^{3}$International Computer Science Institute (ICSI)}
}

\begin{document}

\maketitle
\thispagestyle{empty}
\pagestyle{empty}

\begin{abstract} 
Policy search can in principle acquire complex strategies for control of robots and other autonomous systems. When the policy is trained to process raw sensory inputs, such as images and depth maps, it can also acquire a strategy that combines perception and control. However, effectively processing such complex inputs requires an expressive policy class, such as a large neural network. These high-dimensional policies are difficult to train, especially when learning to control safety-critical systems. We propose \plato, a \NEW{continuous, reset-free reinforcement learning} algorithm that trains complex control policies with supervised learning, using model-predictive control (MPC) to generate the supervision, hence never in need of running a partially trained and potentially unsafe policy. \plato uses an adaptive training method to modify the behavior of MPC to gradually match the learned policy in order to generate training samples at states that are likely to be visited by the learned policy. \plato also maintains the MPC cost as an objective to avoid highly undesirable actions that would result from strictly following the learned policy before it has been fully trained. We prove that this type of adaptive MPC expert produces supervision that leads to good long-horizon performance of the resulting policy.We also empirically demonstrate that MPC can still avoid dangerous on-policy actions in unexpected situations during training. Our empirical results on a set of challenging simulated aerial vehicle tasks demonstrate that, compared to prior methods, \plato learns faster, experiences substantially fewer catastrophic failures (crashes) during training, and often converges to a better policy.
\end{abstract}

\section{Introduction}
\label{sec:intro}

Policy search via optimization or reinforcement learning (RL) holds the promise of automating a wide range of decision making and control tasks, in domains ranging from robotic manipulation to self-driving vehicles. One particularly appealing prospect is to use policy search techniques to automatically acquire policies that subsume perception and control, thereby acquiring end-to-end perception-control systems that are adapted to the task.

However, representing policies that combine perception and control requires either a careful choice of features or the use of expressive function approximators. Recent results in perception domains, such as computer vision, natural language processing, and speech recognition, suggest that expressive function approximators, such as neural networks, can outperform hand-designed features when trained directly on raw input data while requiring substantially less manual engineering~\cite{Lecun2015_Nature}. Recent years have seen considerable research on using deep networks for control \cite{Mnih2013_NIPS,Giusti2016_RAL, ChenICCV_2015,Levine2015_endtoend,Lillicrap2016_ICLR,Heess2015_NIPS}.

Unfortunately, training such large, high-dimensional policies on real physical systems is exceedingly challenging for two reasons. First, standard model-free reinforcement learning algorithms are difficult to apply to large non-linear function approximators \cite{Deisenroth2011}. Several recent methods demonstrate RL-based training of large neural networks \cite{Mnih2013_NIPS,Schulman2015_ICML,Lillicrap2016_ICLR,Heess2015_NIPS}, but these approaches require a very large amount of experience, making them difficult to run on physical systems. In contrast, methods based on supervised learning, including DAgger \cite{Ross2011_AISTATS}, guided policy search \cite{Levine2013_ICML,Levine2015_endtoend} and the work presented in this paper, are more sample-efficient, but require a viable source of supervision. The second obstacle to using RL in the real world is that, although a fully trained neural network controller can be very robust and reliable, a partially trained policy can perform unreasonable and even unsafe actions \cite{Ross2011_AISTATS}. This can be a major problem when the agent is a mobile robot or autonomous vehicle and unsafe actions can cause the destruction of the robot or damage to its surroundings.

\begin{figure}[tb]
  \centering
  \includegraphics[width=0.6\columnwidth]{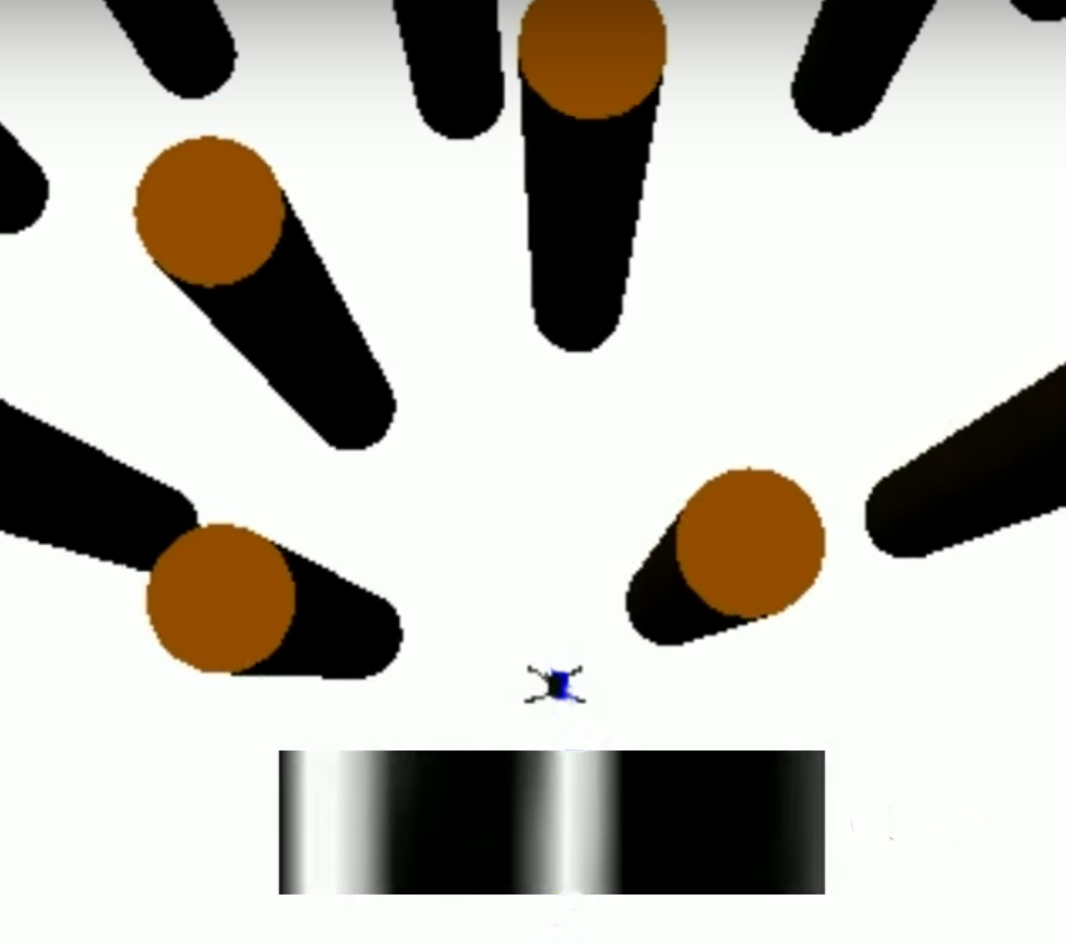}
  \caption{\textbf{Policy Learning using Adaptive Trajectory Optimization}: A neural network control policy trained by PLATO navigates through a forest using camera images. During training, the adaptive MPC teacher policy chooses actions to achieve good long-horizon task performance while matching the learner policy distribution. The policy learned by PLATO converges with bounded cost.}
  \vspace{-20pt}
  \label{fig:simulator}
\end{figure}

We propose \plato (Policy Learning using Adaptive Trajectory Optimization), a \NEW{reset-free} method for training complex policies that combine perception and control by using a trajectory optimization teacher in the form of model-predictive control (MPC). At training time, MPC chooses actions that make a tradeoff between succeeding at the task and matching the behavior of the current policy. By gradually adapting to the policy, MPC ensures that the states visited during training will allow the policy to learn good long-horizon performance. MPC makes use of full state information, which could be obtained, for example, by instrumenting the environment at training time. The final policy, however, is trained to mimic the MPC actions using only the observations available to the robot, which makes it possible to run the resulting policy at test time without any instrumentation. The algorithm requires access to at least a rough model of the system dynamics in order to run MPC during training, but does not require any knowledge of the observation model, making it feasible to use with complex, raw observation signals, such as images and depth scans. Since MPC is used to select all actions at training time, the algorithm never requires running a partially trained and potentially unsafe policy.%

We prove that the policy learned by \plato converges to a policy with bounded cost. Our empirical results further demonstrate that \plato can learn complex policies for simulated quadrotor flight with laser rangefinder observations and camera observations in cluttered environments and at high speeds. We show that \plato outperforms a number of previous approaches in terms of both the performance of the final neural network policy and the robustness to catastrophic failure during training. In comparisons with MPC-guided policy search \cite{Zhang2016_ICRA}, the DAgger algorithm \cite{Ross2011_AISTATS}, DAgger with coaching \cite{He2012_NIPS} and supervised learning, our approach experiences substantially fewer catastrophic failures both during training time and at test time.

\section{Related Work}
\label{related}

Deep neural networks have emerged as powerful general-purpose models for processing complex sensory data. Recent years have seen an increasing amount of research on using neural networks to represent control policies for control tasks, including playing Atari games~\cite{Mnih2013_NIPS}, robotic manipulation from camera images~\cite{Levine2015_endtoend}, and various other continuous control tasks \cite{Schulman2015_ICML, Lillicrap2016_ICLR,Heess2015_NIPS}. Broadly speaking, these methods fall into two categories: methods based on reinforcement learning (RL), including Q-learning~\cite{Mnih2013_NIPS} and policy search~\cite{Schulman2015_ICML,Lillicrap2016_ICLR},
and methods based on supervised learning, including DAgger~\cite{Ross2011_AISTATS} and guided policy search~\cite{Levine2013_ICML}.

RL-based methods are typically more general, but require a very large amount of system experience, which limits their applicability to real physical systems~\cite{Deisenroth2011}. Furthermore, the need to explore using partially learned policies, or worse, using random actions, causes these methods to exhibit potentially dangerous and unstable behavior during training. These limitations make it difficult to deploy RL-based algorithms directly on safety-critical systems.

Methods based on supervised learning are dramatically more sample-efficient, but require a viable source of supervision. In the case of guided policy search, this supervision comes from a simpler RL algorithm that does not directly optimize a neural network policy, but a much simpler trajectory-centric controller~\cite{Levine2014_NIPS}. This approach typically requires the ability to deterministically reset the environment, which is not always feasible when learning in the real world. In the case of DAgger, supervision can come from a human expert~\cite{Ross2013_ICRA} or a computational expert, such as Monte Carlo tree search~\cite{Guo2014_NIPS}. However, this expert does not adapt to the learned neural network policy, and successful application of DAgger assumes that the learned policy can mimic the expert up to a small bounded error~\cite{Ross2011_AISTATS}. This assumption is not always realistic~\cite{Levine13}. Furthermore, DAgger requires executing the learned policy at training time to acquire samples from its own state distribution. When learning is performed online in non-stationary environments, this can expose the agent to dangerous situations for which the learned policy has not yet been fully trained.

In this paper, we propose \plato, an algorithm that trains neural network control policies with supervised learning, using model-predictive control (MPC) to generate the supervision. In contrast to DAgger, \plato adapts the computational ``expert'' (the MPC algorithm) to the learned policy, but does not actually execute the learned policy in the real world until training is completed. We show that this still enforces a bound on the difference between the state distribution of the learned policy and that of the MPC expert, but has the benefit of not exposing the agent to dangerous situations since MPC can still avoid dangerous on-policy actions in novel situations. 

Furthermore, we demonstrate that \plato can be used to train a policy that uses raw perceptual input, while the MPC teacher uses the true state, which allows us to train the policy without access to a model of the sensors, similarly to recent work on guided policy search~\cite{Levine2015_endtoend,Zhang2016_ICRA}. However, \plato lifts a major limitation of guided policy search, which is the requirement to reset the environment between episodes -- in fact, \plato does not even assume an episodic formulation of the task; a practical training scenario might consist, for example, of a robot continuously and autonomously exploring its environment with MPC for the duration of the training period. Since resetting the environment for episodic tasks can be complex, time-consuming or even impossible in the real world, not requiring such resets is a major advantage.

\section{Preliminaries and Overview}
\label{sec:pre}

We address the problem of learning control policies for dynamical systems, such as robots and autonomous vehicles. The system is defined by states $\bx$ and actions $\bu$. The policy must control the system from observations $\bo$, which are in general insufficient for determining the full state $\bx$. The policy is a conditional distribution over actions $\pith(\bu | \bo_t)$, parametrized by $\theta$. At test time, the agent chooses actions according to $\pith(\bu | \bo_t)$ at each time step $t$, and experiences a loss $c(\bx_t, \bu_t)$. We assume without loss of generality that $c(\bx_t, \bu_t)$ is in the interval $[0,1]$. The next state is distributed according to the dynamics $p(\bx_{t+1} | \bx_t, \bu_t)$. The goal is to learn a policy $\pith(\bu | \bo_t)$ that minimizes the total cost $J(\pi) = E_{\pi}\big[\sum_{t=1}^T c(\bx_t, \bu_t)\big]$. We will use $J_t(\pi | \bx_t) = E_{\pi}\big[\sum_{t'=t}^T c(\bx_{t'}, \bu_{t'})|\bx_t\big]$ as shorthand for the expected cost from state $\bx_t$ at time $t$, such that $J(\pi) = E_{\bx_1 \sim p(\bx_1)}\big[J_1(\pi | \bx_1)\big]$.

In this work, we further assume that during training, our algorithm has access to the true underlying states $\bx$. This additional assumption allows us to use simple and efficient model-predictive control (MPC) methods to generate training actions. We do not require knowing the true states $\bx$ at test time, since the learned policy $\pith(\bu | \bo_t)$ only requires observations. This training setup could be implemented in various ways in practice, including instrumenting the training environment (e.g. using motion capture to track a mobile robot) or using more effective hardware at training time (such as a more accurate GPS system), while only having access to cheaper and more practical hardware at test time. While this assumption does introduce some restrictions, we will show that it enables very efficient and relatively safe training, making it an appealing option for safety-critical systems.

We will train the policy $\pith(\bu | \bo_t)$ by mimicking a computational ``teacher,'' rather than attempting to learn the policy directly with reinforcement learning. There are three key advantages to this approach: first, the teacher can exploit the true state $\bx$, while the final policy $\pith$ is only trained on the observations $\bo$; second, we can choose a teacher that will remain safe and stable, avoiding dangerous actions during training; third, we can train the final policy $\pith$ using standard, robust supervised learning algorithms, which will allow us to construct a simple and highly data-efficient algorithm that scales easily to complex, high-dimensional policy parametrization. Specifically, we will use MPC as the teacher. MPC uses the true state $\bx$ and a model of the system dynamics (which we assume to be known in advance, but which in general could also be learned from experience). MPC plans locally optimal trajectories with respect to the dynamics, and by replanning every time step, is able to achieve considerable robustness to unexpected perturbations and model errors~\cite{Mayne2005_automatica}, making it an excellent choice for sample-efficient learning.

\section{Policy Learning using Adaptive Trajectory Optimization}

One na\"{i}ve approach to learn a policy from a computational teacher such as MPC would be to generate a training set with MPC, and then train the policy with supervised learning to maximize the log-likelihood of this dataset. The teacher can safely choose robust, near-optimal trajectories. However, this type of supervision ignores the fact that the state distribution for the teacher and that of the learner are different \cite{Ross2011_AISTATS}. Formally, the distribution of states at test time will not match the distribution at training time, and we therefore cannot expect good long-horizon performance from the learned policy.

In order to overcome this challenge, \plato uses an adaptive MPC teacher that modifies its actions in order to bring the state distribution in the training data closer to that of the learned policy, while still producing robust trajectories and reacting intelligently to unexpected perturbations that cannot be handled by a partially trained policy. To that end, the teacher generates actions at each time step $t$ from a controller obtained by optimizing the following objective:
\begin{multline}
\pilat(\bu | \bx_t, \theta) \leftarrow
\arg\min_\pi J_t(\pi | \bx_t)  \\
+ \lambda \dkl\big(\pi(\bu | \bx_t) || \pi_\theta(\bu | \bo_t)\big),
\label{eqn:obj}
\end{multline}%
where $\lambda$ determines the relative importance of matching the learner $\pi_\theta$ versus optimizing the expected return $J(\cdot)$. Since the teacher uses an MPC algorithm, this objective is reoptimized at each time step to obtain a locally optimal controller for the current state. The only difference from a standard MPC algorithm is the inclusion of the KL-divergence term. The particular MPC algorithm we use is based on iterative LQG (iLQG) \cite{Todorov2005_ACC}, using a maximum entropy variant that produces linear-Gaussian stochastic controllers of the form $\pila(\bu | \bx_t) = \mathcal{N}(\mathbf{K}_t\bx_t + \mathbf{k}_t, \Sigma_t)$ \cite{Levine2013_ICML}. The details of this maximum entropy variant of iLQG may be found in prior work \cite{Todorov2005_ACC,Kappen2012_ML,Levine2014_NIPS}. We describe the details of \plato and its relation to prior methods in Sec. \ref{sec:derivation} and show that \plato produces a good learned policy in Sec. \ref{sec:theory}.

\begin{algorithm}[b]
  \caption{\plato algorithm}
  \label{alg:ours}
\begin{algorithmic}[1]
	\STATE Initialize data $\mathcal{D} \leftarrow \emptyset$
	\FOR{$i=1$ {\bfseries to} $N$}
	    \FOR{$t = 1$ {\bfseries to} $T$}
	    \STATE Optimize $\pilat$ with respect to Equation~(\ref{eqn:obj})$\!\!\!$
	    \STATE Sample $\bu_t \sim \pilat(\bu | \bx_t, \theta)$
	    \STATE Optimize $\pi^*$ with respect to Equation~(\ref{eqn:obj-opt})$\!\!\!$
	    \STATE Sample $\bu_t^* \sim \pi^*(\bu | \bx_t)$
	    \STATE Append $\big(\bo_t, \bu_t^*\big)$ to the dataset $\mathcal{D}$
	    \STATE State evolves $\bx_{t+1} \sim p(\bx_{t+1} | \bx_t, \bu_t)$
	    \ENDFOR
		\STATE Train $\pi_{\theta_{i+1}}$ on $\mathcal{D}$
	\ENDFOR
\end{algorithmic}
\end{algorithm}

\subsection{Algorithm Description}
\label{sec:derivation}

Algorithm~\ref{alg:ours} outlines \plato. We collect training trajectories by choosing actions $\bu_t$ according to an adaptive teacher policy $\pilat(\bu | \bx_t, \theta)$, which is generated by optimizing the objective in Equation \ref{eqn:obj} at each time step via iLQG. We then update the learner policy $\pith(\bu | \bo_t)$ with supervised learning at the observations $\bo_t$ corresponding to the visited states $\bx_t$ to minimize the difference between $\pith(\bu | \bo_t)$ and the locally optimal policy
\begin{equation}
\pi^*(\bu | \bx_t) \leftarrow \arg\min_\pi J(\pi),
\label{eqn:obj-opt}
\end{equation}%
which is also obtained via MPC, but without considering the KL-divergence term. This approach ensures the teacher visits states that are similar to those that would be visited by the learner policy $\pith$, while still providing supervision from a near-optimal policy. Note that the MPC policy is conditioned on the state of the system $\bx_t$, while the learned policy $\pith(\bu | \bo_t)$ is only conditioned on the observations. MPC requires access to at least a rough model of the system dynamics, as well as the system state, in order to robustly choose near-optimal actions. However, by training $\pith$ on the corresponding observations, instead of the true states, $\pith$ can learn to process raw sensory inputs without requiring true state observations, making it possible to run the learned policy with only the raw observations at test time. In the rest of this section, we describe the MPC teacher and the supervised learning procedure in detail.

\textbf{Adaptive MPC teacher}: The teacher's policy $\pilat$ must take reasonable, robust actions while visiting states that are similar to those that would be seen by the learner policy $\pith$. However, we do not know the state distribution of $\pi_\theta$ in advance, since although we have some approximate knowledge of the system dynamics, we do not assume a model of the observation function that produces observations $\bo_t$ from states $\bx_t$, making it impossible to simulate the policy $\pith$ into the future. Instead, we choose the actions at each time step according to an MPC policy $\pilat$ that minimizes the expected long-term sum of costs $J_t(\pilat | \bx_t)$, but only greedily minimizes the KL-divergence against $\pith$ at the current time step $t$, where the observation $\bo_t$ is already available, resulting in the objective in Equation \ref{eqn:obj}. Since MPC reoptimizes the local policy at each time step, this method produces a sequence of policies $\pi_{\lambda}^{1:T}$, each of which is optimized with respect to its long-horizon cost and immediate disagreement with $\pith$.

As discussed previously, our iLQG-based MPC algorithm produces linear-Gaussian local controllers $\pilat(\bu|\bx_t) = \mathcal{N}(\mu_\lambda(\bx_t), \Sigma_t)$ where $\mu_\lambda(\bx_t) = \mathbf{K}_t \bx_t + \mathbf{k}_t$. We will further assume that our learner policy is conditionally Gaussian (but nonlinear), though other parametric distributions are also possible. The policy therefore has the form $\pith(\bu | \bo_t) = \mathcal{N}(\mu_\theta(\bo_t), \Sigma_\pith)$ where $\mu_\theta(\bo_t)$ is the output of a nonlinear function, such as a neural network, and covariance $\Sigma_\pith$ can be either learned or deterministic. Then the MPC objective can be expressed in closed form:
\begin{multline*}
\min_\pi \; J_t(\pi | \bx_t) + \frac{1}{2} \lambda \Big[
\ln \left(\frac{|\Sigma_\pith|}{|\Sigma_t|} \right) + \tr\big(\Sigma_{\pith}^{-1} \Sigma_t\big) \\
+ \big(\mu_\theta(\bo_t) - \mu_\lambda(\bx_t)\big)^\intercal \Sigma_{\pith}^{-1} \big(\mu_\theta(\bo_t) - \mu_\lambda(\bx_t)\big) + \text{const}\Big].
\end{multline*}

The KL-divergence term in this objective is quadratic in $\bu_t$ and linear in the covariance $\Sigma_t$, with an entropy maximization term $-\ln |\Sigma_t|$. This is precisely the objective that is optimized by the maximum entropy variant of iLQG \cite{Levine2014_NIPS}, and optimization requires us only to expand the cost-to-go $J_t$ to second order, which is a standard procedure in iLQG.

\textbf{Training the learner's policy}: We want the learner's policy $\pi_\theta$ to approach the optimal policy $\pi^*(\bu | \bx_t)$. We can estimate a (locally) optimal policy $\pi^*$ at each state $\bx_t$ with iLQG, simply by repeating the optimization at each time step but excluding the KL-divergence term. During the supervised learning phase, we minimize the KL-divergence between the learner $\pith$ and the precomputed near-optimal policies $\pi^*$ at the observations stored in the dataset $\mathcal{D}$:
\begin{align}
\theta \leftarrow \arg\min_\theta \; \sum_{( \bx_t, \bo_t ) \in \mathcal{D}} \dkl\big(\pi_\theta(\bu | \bo_t) || \pi^*(\bu | \bx_t)\big). \label{eq:sl}
\end{align}

Since both $\pith$ and $\pi^*$ are conditionally Gaussian, the KL-divergence can be expressed in closed-form:
\begin{small}
\begin{multline*}
\min_\theta \; \frac{1}{2} \sum_{(\bx_t, \bo_t) \in \mathcal{D}} 
\big(\mu^*(\bx_t) - \mu_\theta(\bo_t)\big)^\intercal \Sigma_{\pi^*}^{-1} \big(\mu^*(\bx_t) - \mu_\theta(\bo_t)\big) \\
+ \tr\big(\Sigma_{\pi^*}^{-1} \Sigma_\pith\big) +  \ln\left(\frac{|\Sigma_{\pi^*}|}{|\Sigma_\pith|}\right) + \text{const}.
\end{multline*}
\end{small}%
Ignoring the terms that do not involve the learner policy mean $\mu_\theta(\bo_t)$, the objective function can be rewritten in the form of a weighted Euclidean loss:
\begin{small}
\begin{align*}
\min_\theta \; \sum_{(\bx_t, \bo_t) \in \mathcal{D}} ||\mu^*(\bx_t) - \mu_\theta(\bo_t)||_{\Sigma_{\pi^*}^{-1/2}}^2.
\end{align*}
\end{small}%
This optimization can then be solved using standard regression methods. In our experiments, $\mu_\theta$ is represented by a neural network, and the above optimization problem corresponds to standard neural network regression, solvable by stochastic gradient descent. The covariance of $\pith$ can be solved for in closed form, and corresponds to the inverse of the average precisions of $\pi^*$ at the training points~\cite{Levine2013_ICML}.

\begin{table}[t]
\vspace*{5pt}
\begin{center}
{
\begin{tabular}{|c|c|c|}
\hline approach & \specialcell{teacher\\policy} & \specialcell{supervision\\policy} \\
\hline supervised learning & $\pi^*$ & $\pi^*$ \\
\hline DAgger & $\pimix$ & $\pi^*$ \\
\hline DAgger + coaching & $\pimix$ & $\picoach$ \\
\hline \hline \plato & $\pila$ & $\pi^*$ \\
\hline 
\end{tabular}
}
\caption{
\textbf{Overview of teacher-based policy optimization methods}: For \plato and each prior approach, we list which teacher policy is used for sampling trajectories and which supervision policy is used for generating training actions from the sampled trajectories. Note that the prior methods execute the mixture policy $\pimix$, which requires running the learned policy $\pith$, potentially executing dangerous actions when $\pith$ is not fully trained.
} 
\label{table:approaches}
\end{center}
\vspace*{-30pt}
\end{table}

\vspace*{-2pt}
\subsection{Relationship to previous work}
\label{sec:relationship}
\vspace*{-2pt}

The motivation behind \plato is most similar to the MPC variant of guided policy search (MPC-GPS)~\cite{Zhang2016_ICRA}. However, \plato lifts a major limitation of MPC-GPS. MPC-GPS requires the ability to deterministically reset the environment into one of a small set of initial states. \NEW{MPC-GPS requires deterministic resets because the KL-divergence term is evaluated using a linearization around each rollout.}  Deterministic episodic resets can be complex, time-consuming, or even impossible in the real world. For example, imagine a robot learning to navigate a human crowd; deterministic resets would require having the crowd walk through the same paths in each episode. Not requiring such resets is a major advantage. Furthermore, even when deterministic resets are feasible, \plato empirically outperforms MPC-GPS (Section~\ref{sec:exp}).

Formally, \plato can also be viewed as a generalization of the Dataset Aggregation (DAgger) algorithm \cite{Ross2011_AISTATS}, which samples trajectories according to the mixture policy $\pimix_i = \beta_i \pi^* + (1-\beta_i) \pith_i$. The training data is generated from the observations sampled by executing $\pimix_i$ but labelled with actions from $\pi^*$. DAgger converges if ${\frac{1}{N}\sum_{i=1}^N \beta_i \rightarrow 0}$ as $N \rightarrow \infty$. Coaching~\cite{He2012_NIPS}, a related extension to DAgger, modifies the supervision policy $\pi^*$ to adapt to the learned policy $\pith$ by labelling the training data with a coach policy $\picoach$ that encourages the action training labels to be similar to the actions $\pith_i$ would choose. Our empirical evaluation shows that \plato outperforms coaching.

Another distinction of \plato is the use of an adaptive MPC policy $\pi_{\lambda}^{1:T}$ to select the actions at each time step, rather than the mixture policy $\pimix$ used in the prior methods. As demonstrated in our evaluation, this adaptive MPC policy allows \plato to robustly avoid catastrophic failure during training, which is particularly important in safety-critical domains. Our experiments also demonstrate that policies trained using \plato empirically outperform policies trained by either DAgger or coaching. Table \ref{table:approaches} summarizes the teacher and supervision policies used by \plato and prior work.

\vspace*{20pt}

\section{Theoretical Analysis}
\label{sec:theory}

In this section, we present a proof that the policy $\pith$ learned by \plato converges to a policy with bounded cost. This proof extends the result by Ross et al.~\cite{Ross2011_AISTATS}, which only admits mixture policies, to our adaptive MPC policy $\pi_{\lambda}^{1:T}$.

Given a policy $\pi$, we denote $d_{\pi}^t$ as the state distribution at time $t$ when executing policy $\pi$ from time $1$ to $t-1$. Define the cost function $c(\bx_t, \bu_t)$ as a function of state $\bx_t$ and control $\bu_t$, with $c(\bx_t, \bu_t) \in [0,1]$ without loss of generality. We wish to learn a policy $\pith(\bu | \bo_t)$ that minimizes the total expected cost over time horizon $T$:
\begin{align*}
J(\pi) = \sum_{t=1}^T \E_{\bx_t \sim d_{\pith}^t} [\E_{\bu_t \sim \pith(\bu | \bo_t)} [ c(\bx_t, \bu_t) | \bx_t]].
\end{align*}

Let $J_t(\pi, \tilde{\pi})$ denote the expected cost of executing $\pi$ for $t$ time steps, and then executing $\tilde{\pi}$ for the remaining $T-t$ time steps, and let $Q_t(\bx, \pi, \tilde{\pi})$ denote the cost of executing $\pi$ for one time step starting from initial state $\bx$, and then executing $\tilde{\pi}$ for the remaining $t-1$ time steps. We assume the cost-to-go difference between the learned policy and the optimal policy is bounded: $Q_t(\bx, \pith, \pi^*) - Q_t(\bx, \pi^*, \pi^*) \leq \delta$. In the worst case, $\delta$ is $O(T)$ and \plato (as well as similar methods such as DAgger) will not outperform supervised learning. However, if $\pi^*$ is able to quickly recover from mistakes made by $\pith$, $\delta$ will be $O(1)$~\cite{Ross2011_AISTATS}.

When optimizing Equation \ref{eqn:obj} to obtain the teacher policy $\pila$, we choose $\lambda$ such that $\dkl(\pila(\bu | \bx) || \pith(\bu | \bo)) \leq \epslt$ for all state-observation pairs $(\bx, \bo)$. We can always guarantee this bound when optimizing Equation \ref{eqn:obj}  because $\dkl(\pila(\bu | \bx) || \pith(\bu | \bo)) \rightarrow 0$ as $\lambda \rightarrow \infty$.

When optimizing the supervised learning objective in Equation \ref{eq:sl} to obtain the learner policy $\pith$, we assume the supervised learning objective function error is bounded by a constant $\dkl(\pith(\bu | \bo) || \pi^*(\bu | \bx)) \leq \epsts$ for all states $\bx$ (and corresponding observations $\bo$) in the dataset, which were sampled from the teacher policy distribution $d_\pila$. Since the policy $\pith$ is trained with supervised learning precisely on these states $\bx \sim d_\pila$, this bound $\epsts$ corresponds to assuming that the learner policy $\pith$ attains bounded training error.

Let $l(\bx, \pith, \pi^*)$ denote the expected 0-1 loss of $\pith$ with respect to $\pi^*$ in state $\bx$: $\E_{\bu_\theta \sim \pith(\bu | \bo), \bu^* \sim \pi^*(\bu | \bx)} [ \mathbf{1}[\bu_\theta \neq \bu^*]]$. We note that the total variation divergence is an upper bound on the 0-1 loss  \cite{Nguyen2005_NIPS} and the KL-divergence is an upper bound on the total variation divergence \cite{Pollard2000}. Therefore for all states $\bx \sim d_\pila$ in the dataset used for supervised learning, the 0-1 loss can be upper bounded:
\begin{align*}
l(\bx, \pith, \pi^*) &= \E_{\bu_\theta \sim \pith(\bu | \bo), \bu^* \sim \pi^*(\bu | \bx)} [ \mathbf{1}[\bu_\theta \neq \bu^*]] \\
&\leq \dtv(\pith(\bu | \bo) || \pi^*(\bu | \bx)) \\
&\leq \sqrt{\dkl(\pith(\bu | \bo) || \pi^*(\bu | \bx))} \\
&\leq \sqrt{\epsts}.
\end{align*}

We also note the state distribution bound $||d_\pi^t - d_{\tilde{\pi}}^t||_1 \leq 2 t \sqrt{\dklmax(\pi, \tilde{\pi})}$ proven in \cite{Schulman2015_ICML}. This lemma implies that for an arbitrary function $f(\bx)$, $E_{\bx \sim d_\pi^t}[f(\bx)] \leq E_{\bx \sim d_{\tilde{\pi}}^t}[f(\bx)] + 2 f^{\max} t \sqrt{\dklmax(\pi, \tilde{\pi})}$

We can then prove the following theorem:

\begin{theorem}
Let \NEW{the cost-to-go} $Q_t(\bx, \pith, \pi^*) - Q_t(\bx, \pi^*, \pi^*) \leq \delta$ for all $t \in \{1, ..., T\}$ . Then for \textnormal{\plato}, $J(\pith) \leq J(\pi^*) + \delta \sqrt{\epsts} O(T) + O(1)$.
\label{theorem:ours}
\end{theorem}

\noindent \textit{Proof}:
{\small
\begin{subequations}
\begin{align}
\!\!\!\!J(\pith)
&= J(\pi^*) + \sum_{t=0}^{T-1} J_{t+1}(\pith, \pi^*) - J_{t}(\pith, \pi^*) \nonumber \\
&= J(\pi^*) + \sum_{t=1}^T \E_{\bx \sim d_{\pith}^t}[Q_{t}(\bx, \pith, \pi^*) - Q_{t}(\bx, \pi^*, \pi^*)] \nonumber \\
&\leq J(\pi^*) + \delta \sum_{t=1}^T \E_{\bx \sim d_{\pith}^t} [ l(\bx, \pith, \pi^*) ] \label{eq:proof-loss} \\
&\leq J(\pi^*) + \delta \sum_{t=1}^T \E_{\bx \sim d_{\pila}^t} [ l(\bx, \pith, \pi^*) ] \!+\! 2 l^{\max} t \sqrt{\epslt} \label{eq:proof-dist} \\
&\leq J(\pi^*) + \delta \sum_{t=1}^T \sqrt{\epsts} + 2 t \sqrt{\epsts} \sqrt{\epslt} \label{eq:proof-loss-upper} \\
&= J(\pi^*) + \delta T \sqrt{\epsts} + \delta T (T+1) \sqrt{\epsts} \sqrt{\epslt} \nonumber
\end{align}
\end{subequations}
}%
Equation \ref{eq:proof-loss} follows from the fact that the expected 0-1 loss of $\pith$ with respect to $\pi^*$ is the probability that $\pith$ and $\pi^*$ pick different actions in $\bx$; when they choose different actions, the cost-to-go increases by $\leq \delta$. Equation \ref{eq:proof-dist} follows from the state distribution bound proven in \cite{Schulman2015_ICML}. Equation \ref{eq:proof-loss-upper} follows from the upper bound on the 0-1 loss.

Although we do not get to choose $\epsts$ because that is a property of the supervised learning algorithm and the data, we are able to choose $\epslt$ by varying parameter $\lambda$. If we choose $\lambda$ such that $\epslt = O(\frac{1}{T^2})$. We therefore have
\begin{align*}
J(\pith) \leq J(\pi^*) + \delta \sqrt{\epsts} O(T) + O(1) \hspace{0.1in} \tag*{$\square$}.
\end{align*}

As with DAgger, in the worst case $\delta = O(T)$. However, in many cases $\delta = O(1)$ or is sub-linear in $T$, for instance if $\pi^*$ is able to quickly recover from mistakes made by $\pith$.
We also note that this bound, $O(T)$, is the same as the bound obtained by DAgger, but without actually needing to directly execute $\pith$ at training time. Compared to supervised learning with bound $O(T^2)$ \cite{Ross2011_AISTATS}, \plato trains the policy at states closer to those induced under its own distribution.

\section{Experiments}
\label{sec:exp}

\begin{figure*}[t]
\vspace*{0pt}
\captionsetup[subfigure]{aboveskip=-0.2em,belowskip=-0.2em}
\centering
\begin{subfigure}[b]{0.65\textwidth}
\includegraphics[width=\textwidth,trim=5em 4em 5em 0em,clip]{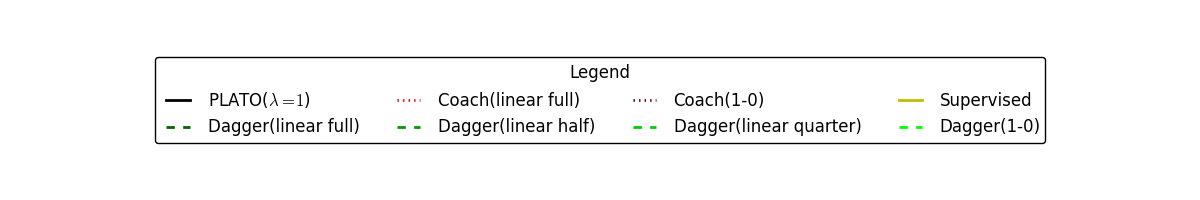}
\end{subfigure}
\begin{subfigure}[t]{0.32\textwidth}
\includegraphics[width=\textwidth,trim=1em 0 5em 2em,clip]{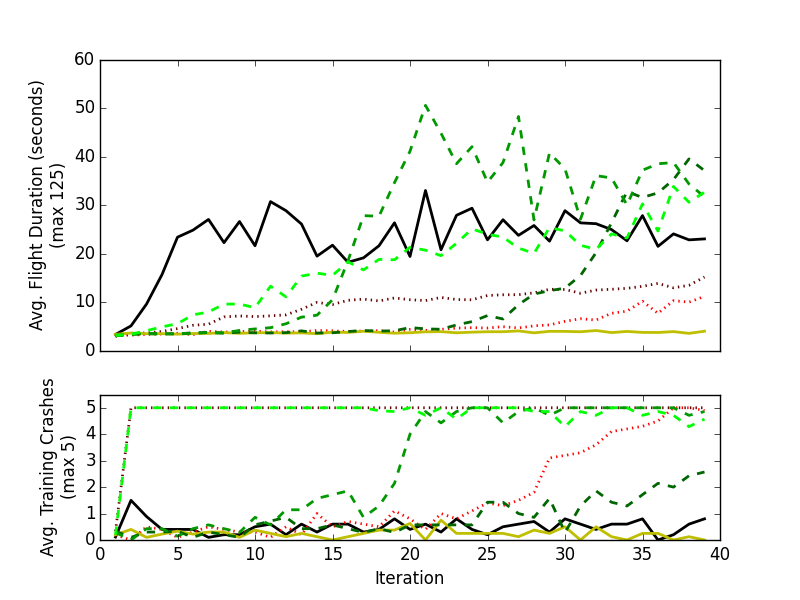}
\caption{canyon (laser)}
\label{fig:exp-hallway-laser}
\end{subfigure}~
\begin{subfigure}[t]{0.32\textwidth}
\includegraphics[width=\textwidth,trim=1em 0 5em 2em,clip]{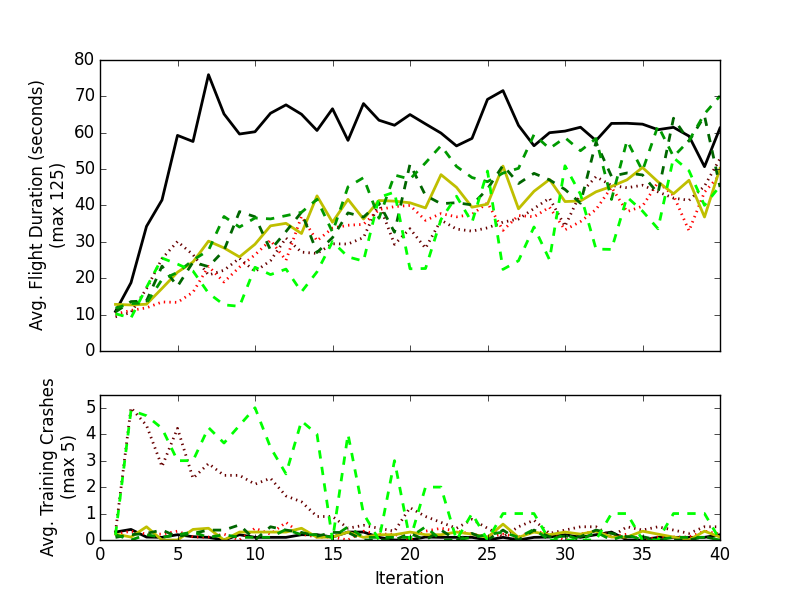}
\caption{canyon (camera)}
\label{fig:exp-hallway-camera}
\end{subfigure}~
\begin{subfigure}[t]{0.32\textwidth}
\includegraphics[width=\textwidth,trim=1em 0 5em 2em,clip]{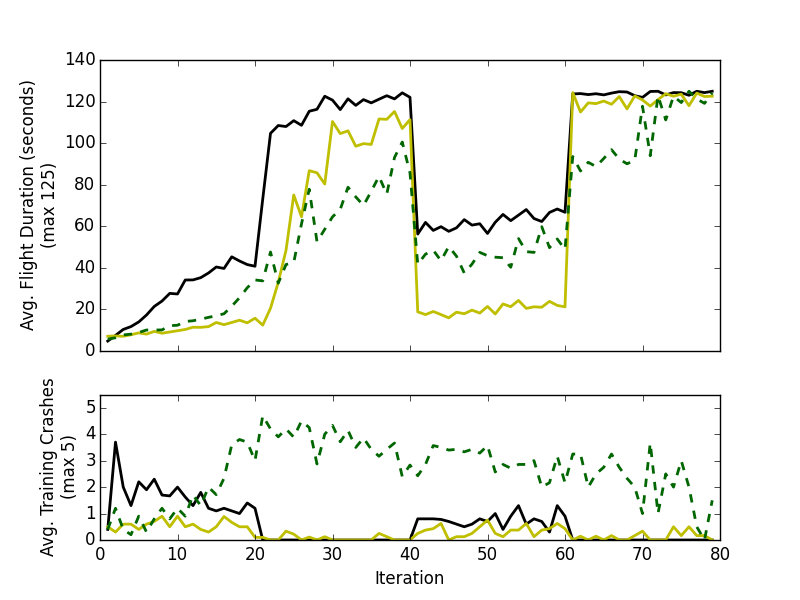}
\caption{canyon/forest switching (camera)}
\label{fig:exp-hallway-forest-camera}
\end{subfigure}
\begin{subfigure}[t]{0.32\textwidth}
\includegraphics[width=\textwidth,trim=1em 0 5em 2em,clip]{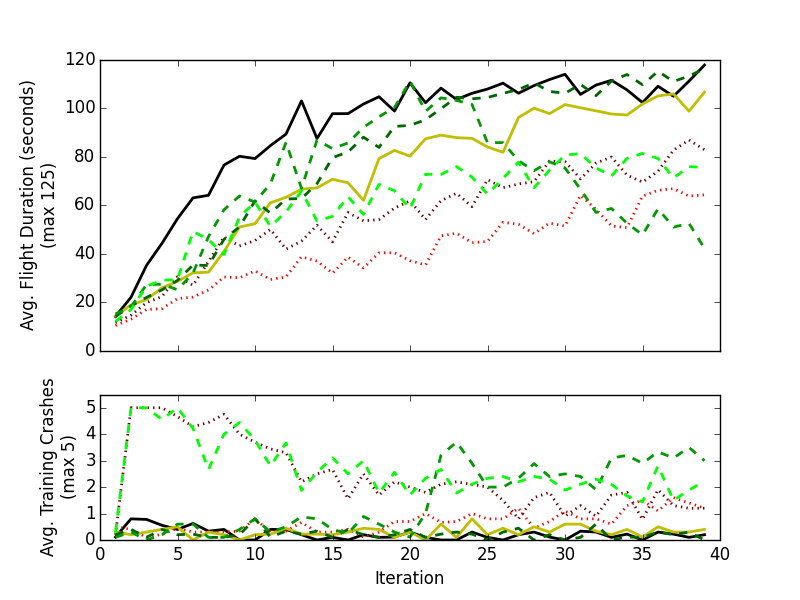}
\caption{forest (laser)}
\label{fig:exp-forest-laser}
\end{subfigure}~
\begin{subfigure}[t]{0.32\textwidth}
\includegraphics[width=\textwidth,trim=1em 0 5em 2em,clip]{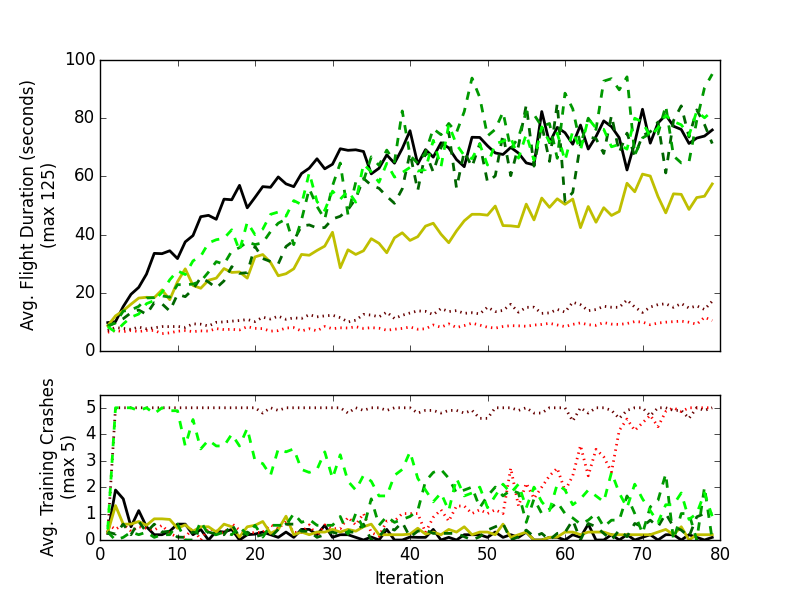}
\caption{forest (camera)}
\label{fig:exp-forest-camera}
\end{subfigure}~
\begin{subfigure}[t]{0.32\textwidth}
\includegraphics[width=\textwidth,trim=1em 0 5em 2em,clip]{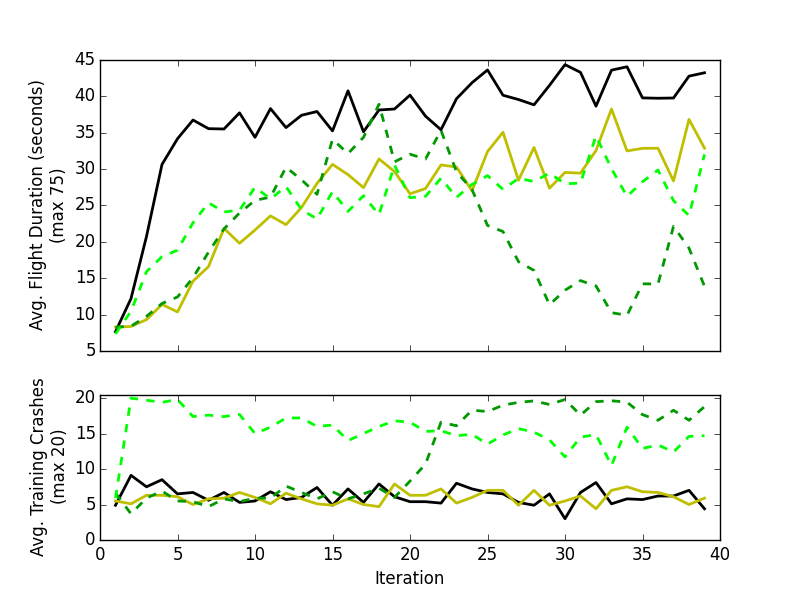}
\caption{velocity commands in forest (laser)}
\label{fig:exp-forest-laser-param}
\end{subfigure}
\caption{\textbf{Experiments}: We compare \plato to baseline methods in a winding canyon, a dense forest, and an alternating canyon/forest. For each scenario and learning method, we trained 10 different policies using different random seeds. Each iteration required 2 minutes of flight time. Then for each policy, we evaluated the neural network policy trained at each iteration by flying through the scenario 20 times. Therefore each datapoint corresponds to 200 samples.}
\label{fig:experiments}
\vspace*{-20pt}
\end{figure*}

We evaluate \plato on a series of simulated quadrotor navigation tasks. MPC is a standard choice for quadrotor control~\cite{Mueller2013_ECC} because approximate models are typically known in advance from standard rigid body physics and the vehicle specifications. However, effective use of MPC requires explicit state estimation and can be computationally intensive. It is therefore very appealing to be able to train an entirely feedforward, reactive policy to control a quadrotor performing navigation in obstacle-rich environments, directly in response to raw sensor inputs. During training, the vehicle might be placed in a known, instrumented training environment to collect data using MPC, while at test time, the learned feedforward policy could control the aircraft directly from raw observations. This makes simulated quadrotor navigation an ideal domain in which to compare \plato to prior work. 

\begin{figure}[b]
\vspace*{-20pt}
\captionsetup[subfigure]{aboveskip=-0.0em,belowskip=-0.0em}
\centering
\begin{subfigure}[t]{0.51\columnwidth}
\includegraphics[width=\columnwidth]{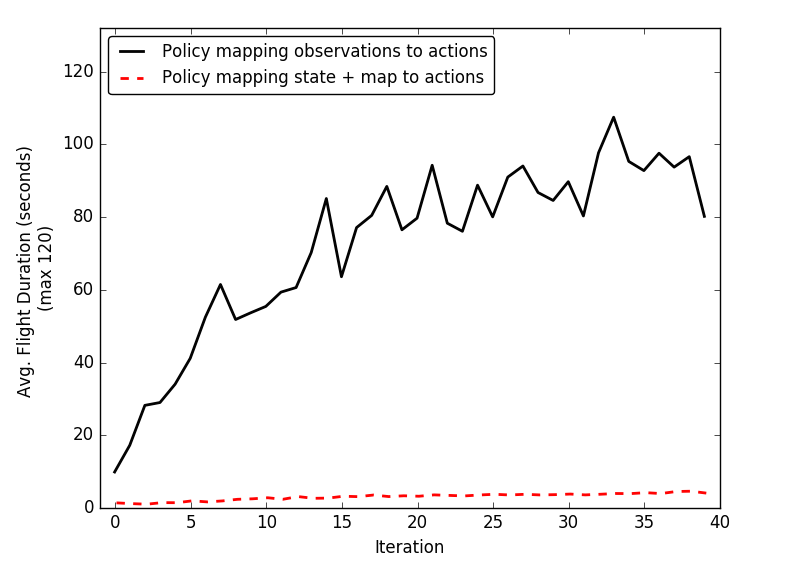}
\caption{}
\label{fig:exp-state-to-action}
\end{subfigure}~
\begin{subfigure}[t]{0.47\columnwidth}
\includegraphics[width=\columnwidth]{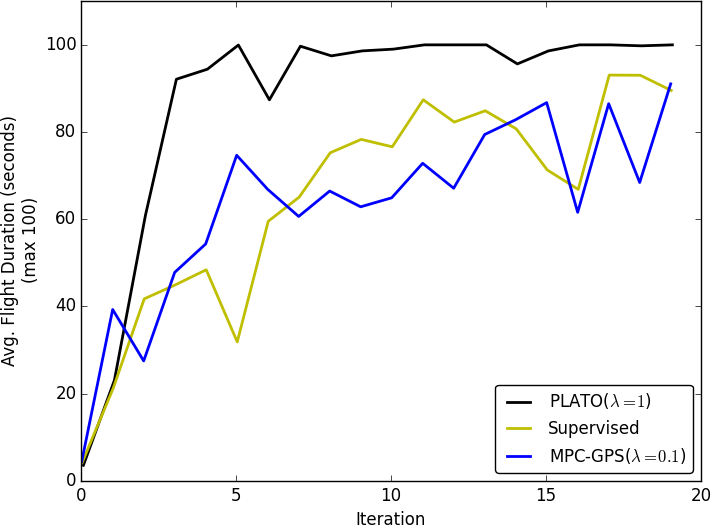}
\caption{}
\label{fig:mpcgps-laser-forest}
\end{subfigure}~
\caption{Comparisons with (a) training on full state and (b) MPC-GPS.}
\label{fig:exp-full-state-and-mpc-gps}
\end{figure}

\textbf{Prior methods and baselines}: We compare \plato to four methods. %
The first method is DAgger, which, as discussed in Section~\ref{sec:relationship}, executes a mixture of the learned policy and teacher policy, which in this case is MPC (without a KL-divergence term). DAgger has previously been used for learning quadrotor control policies from human demonstrations \cite{Ross2013_ICRA}. While DAgger carries the same convergence guarantees as \plato, successful use of DAgger requires the learned policy to be executed at training time, before the policy has converged to a near-optimal behavior. The second method is the coaching algorithm of \cite{He2012_NIPS} which, like DAgger, executes a mixture of the learned and teacher policies, but supervises the learner using the adapted policy. In these experiments, we chose the coaching policy $\picoach$ to be the teacher policy $\pila$ from \plato. For both DAgger and coaching, we must choose the mixing parameter $\beta_i$ at each iteration $i$. Since the performance of these algorithms is quite sensitive to the schedule of the $\beta_i$ parameter, we include four schedules for comparison: three linear schedules that interpolate $\beta_i$ from $1$ at the first iteration to $0$ at the last iteration (``linear full''), the halfway iteration (``linear half''), and the quarter-way iteration (``linear quarter''), as well as the more standard ``1-0'' schedule that sets $\beta_i = 1[i=1]$. The third method is MPC-GPS \cite{Zhang2016_ICRA}, which, unlike \plato, DAgger and coaching, requires deterministic resets during training (Figure~\ref{fig:mpcgps-laser-forest}). In addition to these prior methods, we also compare our approach to a standard supervised learning baseline, which always executes the MPC policy without any adaptation. For all experiments, we assume additive Gaussian noise is applied to both controls and observations.

\textbf{Policy representation}: For all of the methods, we represent $\pith$ as a conditional Gaussian policy, with a constant covariance and a mean given by a neural network function of the observation $\bo_t$. The network has two fully connected hidden layers of size 40 with ReLU activations~\cite{Nair2010_ICML}. The loss function is the weighted euclidean loss (see Section \ref{sec:derivation}). We used the Caffe~\cite{Jia2014_caffe} framework and the ADAM solver~\cite{Kingma2015_ICLR}. Each iteration was trained using the final weights from the previous iteration.

\textbf{Experimental domains}: The comparisons are conducted on two test environments: a winding canyon with randomized turns, and a dense forest of cylindrical trees with randomized positions. An example environment is shown in Figure \ref{fig:simulator}. The canyon changes direction up to $\frac{\pi}{4}$ radians every 0.5m. The forest is composed of 0.5m radius cylinders with an average spacing of 2.5m. The target velocity is 6m/s in the canyon and 2m/s in the forest.

The dynamical system is a quadrotor with dynamics described by \cite{Martin2010_ICRA}. The state of the vehicle $\bx \in \mathbb{R}^{13}$ consists of the position and orientation, as well as their time derivatives, and the control $\bu \in \mathbb{R}^4$ consists of motor velocities. The observations $\bo$ consist of orientation, linear velocity, angular velocity and either (i) a set of 30 equally spaced 1-d laser depth scanners arranged in 180 degree fan in front of the vehicle ($\bo \in \mathbb{R}^{40}$) or (ii) a $5 \times 20$ grayscale camera image ($\bo \in \mathbb{R}^{110}$). Learning neural network policies with these observations forces the policies to perform both perception and control, since success on each of the domains requires avoiding obstacles using only raw sensory input.

The cost function for the MPC teacher encourages the quadrotor to fly at a specific linear velocity and orientation while minimizing control effort and avoiding collisions:
\begin{small}
\begin{align*}
L(\bx, \bu) = &10^3 ||\bx_{\textsc{linvel}} - \bx^*_{\textsc{linvel}}||_2^2 + 10^3||\bx_{\textsc{height}} - \bx^*_{\textsc{height}}||_2^2 + \\
&10^4 ||\bx_{\textsc{quat}} - \bx^*_{\textsc{quat}}||_2^2 + 250 ||\bx_{\textsc{angvel}}||_2^2 + \\
&5^{-3}||\bu - \bu_{\textsc{hover}}||_2^2 + \\
&10^3 \max(d_{\textsc{safe}} - \text{signed-distance}(\bx), 0),
\end{align*}
\end{small}%
where $\bx_{\textsc{linvel}}, \bx_{\textsc{height}}, \bx_{\textsc{quat}}, \bx_{\textsc{angvel}}$ are the linear velocity, height, orientation, and angular velocity of the state $\bx$, respectively; $\bx^*_{\textsc{linvel}}, \bx^*_{\textsc{height}}, \bx^*_{\textsc{quat}}$ are the target linear velocity, height, and orientation, respectively; and $\bu_{\textsc{hover}}$ is the rotor velocity when the quadrotor is hovering. The final term is a hinge loss on the distance of the quadrotor to the nearest obstacle; there is no penalty if the nearest obstacle is further than $d_{\textsc{safe}}$.

\textbf{Performance of learned policies}: In Figures~\ref{fig:exp-hallway-laser}, \ref{fig:exp-hallway-camera}, \ref{fig:exp-forest-laser}, and \ref{fig:exp-forest-camera}, we present the mean time to failure (MTTF) of the learned policy $\pith$ on the canyon and forest environments using the laser or camera sensors. The graphs show the MTTF of each policy at each iteration of the learning process, averaged over 10 training runs of each method with 20 repetitions each. Failure occurs when the quadrotor crashes into an obstacle, with the maximum flight time for each domain listed on the graphs. The results indicate that the \plato algorithm is able to learn effective policies faster, and converges to a solution that is better than or comparable to the baseline methods. For some choices of $\beta$ schedule and supervision scheme, some DAgger variants achieve similar final MTTFs, but always at a slower rate and, as discussed next, with significantly more training crashes.

\textbf{Robustness during training}: In Figures~\ref{fig:exp-hallway-laser}, \ref{fig:exp-hallway-camera}, \ref{fig:exp-forest-laser}, and \ref{fig:exp-forest-camera}, we show the number of crashes experienced during training at each iteration. \plato on average experiences less than one crash per iteration, comparable in performance to the baseline MPC method (supervised learning), indicating that mimicking the learner with a KL-divergence penalty does not substantially degrade the robustness of MPC. In contrast, both DAgger and coaching begin to experience a substantial number of failures when the mixing constant $\beta$ drops. By carefully selecting the schedule for $\beta$, the number of crashes can be reduced.

However, even with a carefully chosen schedule, the prior methods are vulnerable to non-stationary training environments, as illustrated in Figure~\ref{fig:exp-hallway-forest-camera}. In this experiment, the vehicle switches from the canyon to the forest halfway through training, and then switches back to the forest. \NEW{Prior methods that directly execute $\pith$ during training experience many crashes because a policy trained only on the canyon cannot succeed on the forest without additional training. However, \plato experiences on average less than one crash per episode because \plato is able to automatically switch to more off-policy behavior when encountering novel scenarios.} While this example might appear pathological, it is in fact a plausible training setup for a real quadrotor exploring a varied environment, such as different floors of a building. If the walls on one floor are painted, e.g., a different color than the rest, the learned policy could easily experience a catastrophic failure when entering the floor for the first time, even if it was consistently successful on preceding floors.

\textbf{Policies with user velocity commands}: Figure~\ref{fig:exp-forest-laser-param} shows the performance of \plato when learning policies that take an additional input to simulate high-level user control in the form of the desired velocity of the quadrotor. These policies are useful because instead of training multiple policies for different target velocities, we can train one generalizable policy. This input modifies the cost function used by MPC, producing command-aware supervision. During training, the commands vary in the range of $\pm 1$ m/s sideways and $1$ to $2.5$ m/s forward. At test time, we sample velocity commands uniformly at random; the velocity commands are re-sampled whenever the quadrotor reaches the current sampled velocity. The results indicate that \plato can successfully learn such policies, outperforming prior methods and again minimizing the number of crashes during training.

\textbf{Sensitivity to KL-divergence weight}: 
Recall that $\lambda$ determines the degree to which MPC prioritizes following the learner $\pith$ versus performing the desired task. 
As $\lambda \rightarrow 0$, \plato approaches standard supervised learning and is thus safe, while as $\lambda \rightarrow \infty$, \plato approaches DAgger 0-1.
\begin{wrapfigure}{r}{0.55\columnwidth}
  \vspace*{-10pt}
  \centering
  \includegraphics[width=\linewidth]{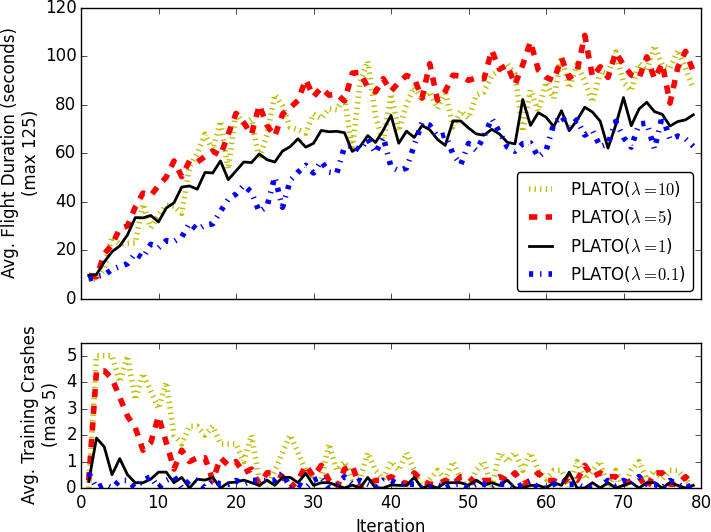}
  \caption{Effect of KL-divergence weight $\lambda$.}
  \label{fig:exp-forest-kl-camera}
  \vspace*{-15pt}
\end{wrapfigure}%
In practice, to choose lambda, we start with $\lambda = 0$, and then increase $\lambda$ until the cost of the behavior starts to increase. Figure~\ref{fig:exp-forest-kl-camera} compares different non-limiting settings of $\lambda$, while we refer to Figure \ref{fig:experiments} for limiting cases for $\lambda=0$ and $\lambda=\infty$. The results suggest that a relatively broad range of $\lambda$ values produces successful policies. 

\textbf{Comparison with training on full state}:
Figure~\ref{fig:exp-state-to-action} shows a comparison where the policy maps state to action using an oracle SLAM algorithm that provides perfect state information and a local 2D distance map of the obstacles. The observation-based policy substantially outperforms the policy that learns to map the state to action, even with an oracle SLAM algorithm. Although the state and obstacle map are sufficient to choose good actions, this mapping is much harder to learn. Of course, alternative full state representations that are carefully engineered to the task may perform better, but this experiment demonstrates that, at least in some cases, mapping observations directly to actions without going through full state estimation can lead to better performance.

\textbf{Comparison with MPC-GPS}: MPC-GPS \cite{Zhang2016_ICRA} cannot directly be evaluated on the domains described above, because training must occur in episodes with deterministic resets (see Section \ref{sec:relationship}). We constructed a fixed-length episodic variant of the forest task where MPC-GPS was allowed to use deterministic resets. Besides not requiring an episodic formulation or deterministic resets, the comparison in Figure~\ref{fig:mpcgps-laser-forest} shows that \plato substantially outperforms the policy learned by MPC-GPS in terms of MTTF.

Supplementary material, \NEW{including a video}, can be viewed online: \url{sites.google.com/site/platopolicy}.

\vspace*{-5pt}
\section{Discussion}
\label{sec:discussion}
\vspace*{-5pt}

In this paper, we presented \plato, a continuous, reset-free algorithm for learning complex, high-dimensional policies that combine perception and control into a single expressive function approximator, such as a deep neural network. \plato uses a trajectory optimization teacher to provide supervision to a standard supervised learning algorithm, allowing for a simple and data-efficient learning method. The teacher adapts to the behavior of the neural network policy to ensure that the distribution over states and observations is sufficiently close to the learned policy, allowing for a bound on the long-term performance of the learned policy. Our empirical evaluation on a simulated quadrotor demonstrates that \plato outperforms a number of previous methods, both in terms of the robustness and effectiveness of the final policy, and in terms of the safety of the training procedure.

\plato has two key advantages that make it well-suited for learning control policies for real-world robotic systems. First, since the learned neural network policy does not need to be executed at training time, the method benefits from the robustness of model-predictive control (MPC), minimizing catastrophic failures at training time. This is particularly important when the training state distribution is non-stationary. Methods that execute the learned policy, such as DAgger, can suffer a catastrophic failure when the agent encounters novel observations. Mitigating these issues typically requires hand-designed safety mechanisms, while \plato automatically switches to a more off-policy behavior.

The second advantage of \plato is that the learned policy can use a different set of observations than MPC. Effective use of MPC requires observing or inferring the full state of the system, which might be accomplished, for instance, by instrumenting the environment with motion capture, or using a known map with relocalization \cite{Williams2007_ICCV}. The policy, however, can be trained directly on raw input from onboard sensors, forcing it to perform both perception and control. Once trained, such a policy can be used in uninstrumented natural environments.

One of the most appealing prospects of learning expressive neural network policies is the possibility of acquiring real-world policies that directly use rich sensory inputs. Because of this, one very interesting avenue for future work is to apply \plato on real physical platforms.

\vspace*{-5pt}
\section{Acknowledgements}
\vspace{-3pt}

This research was funded in part by the Army Research Office through the MAST program, by Darpa under Award \#N66001-15-2-4047, and by Berkeley Deep Drive.

\vspace*{-10pt}

\bibliographystyle{IEEEtran}
\bibliography{2017_ICRA_PLATO}

\end{document}